\pdfoutput=1

\documentclass[11pt]{article}

\usepackage{acl}

\usepackage{times}
\usepackage{latexsym}

\usepackage[T1]{fontenc}

\usepackage[utf8]{inputenc}

\usepackage{microtype}

\usepackage{graphicx} 
\graphicspath{ {./images/} }
\usepackage{multirow}
\usepackage{tabularx}
\usepackage{mathtools}
\usepackage{amsthm}
\usepackage{amsmath}
\usepackage{multirow}
\usepackage{microtype}
\usepackage[rightcaption]{sidecap}
\usepackage{wrapfig}
\usepackage{array}
\usepackage{wrapfig}
\usepackage{varwidth}
\usepackage{hyperref}

\title{ANNA: Enhanced Language Representation for Question Answering}

\author{Changwook Jun, {\bf Hansol Jang}, {\bf Myoseop Sim}, {\bf Hyun Kim}, {\bf Jooyoung Choi}, \\  {\bf Kyungkoo Min} \and {\bf Kyunghoon Bae} \\
        LG AI Research \\
$\left\{\begin{varwidth}{14cm}\centering  cwjun, hansol.jang, myoseop.sim, hyun101.kim, jooyoung.choi, mingk24, k.bae\end{varwidth}\right\}$@lgresearch.ai\\}

\begin{document}
\maketitle
\begin{abstract}
Pre-trained language models have brought significant improvements in performance in a variety of natural language processing tasks. Most existing models performing state-of-the-art results have shown their approaches in the separate perspectives of data processing, pre-training tasks, neural network modeling, or fine-tuning. In this paper, we demonstrate how the approaches affect performance individually, and that the language model performs the best results on a specific question answering task when those approaches are jointly considered in pre-training models. In particular, we propose an extended pre-training task, and a new neighbor-aware mechanism that attends neighboring tokens more to capture the richness of context for pre-training language modeling. Our best model achieves new state-of-the-art results of 95.7\% F1 and 90.6\% EM on SQuAD 1.1 and also outperforms existing pre-trained language models such as RoBERTa, ALBERT, ELECTRA, and XLNet on the SQuAD 2.0 benchmark.    
\end{abstract}

\section{Introduction}

Question answering (QA) is the task of answering given questions, which demands a high level of language understanding and machine reading comprehension abilities. As pre-trained language models based on a transformer encoder~\citep{vaswani2017attention} have brought a huge improvement in performance on a broad range of natural language processing (NLP) tasks including QA tasks, methodologies for QA tasks are widely used to develop applications such as dialog systems~\citep{bansal2021neural} and chat-bots~\citep{hemant2022effect,duggirala2021ita} in a variety of domains.  

\begin{figure}[ht]
\centering
\includegraphics[width=8cm, height=7cm]{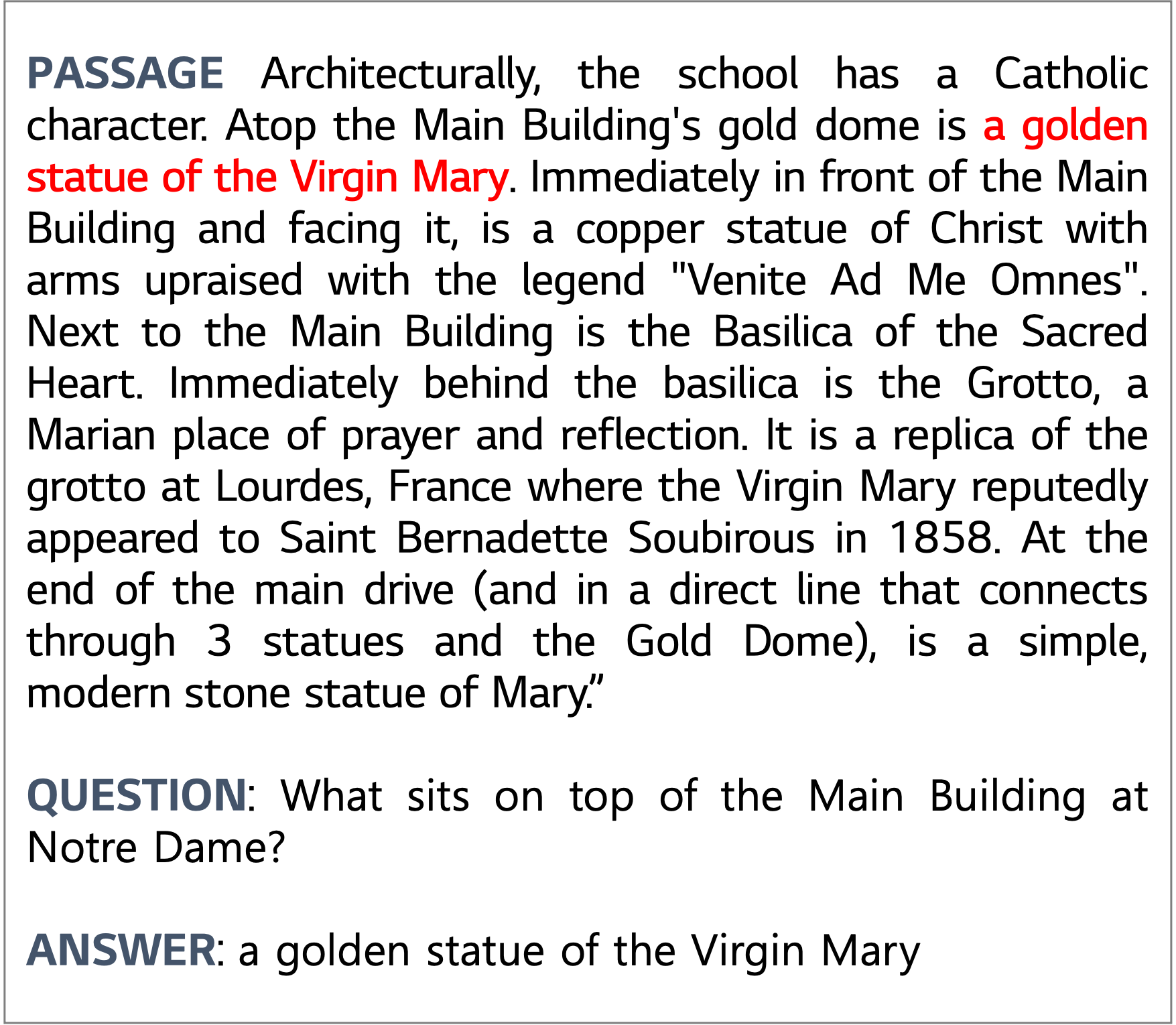}
\caption{Example of a passage with a pair of question and answer sampled from the SQuAD 1.1 dataset.}
\label{fig:fig0}
\end{figure}
Pre-trained language models like BERT~\citep{devlin2018bert} are designed to represent individual words for contextualization. However, recent extractive QA tasks such as Stanford Question Answering Dataset (SQuAD) benchmarks~\citep{rajpurkar2016squad,rajpurkar2018know} involve reasoning relationships between spans of texts that include a group of two or more words in the evidence document~\citep{lee2016learning}. In the example, as shown in Figure~\ref{fig:fig0}, ``a golden statue of the Virgin Mar'', the correct answer for the question ``What sits on top of the Main Building at Notre Dame?'', is a group of words consisting of nouns and other words and is called as a noun phrase, which performs as a noun in a sentence. Since predicting a span of answer texts including a start and end positions may be challenging for self-supervised training rather than predicting an individual word, we introduce a novel pre-training approach that extends a standard masking scheme to wider spans of texts such as a noun-phrase rather than an entity level and prove that this approach is more effective for an extractive QA task by outperforming existing models.

In this paper, we present a new pre-training approach, \textbf{ANNA} (\textbf{A}pproach of \textbf{N}oun-phrase based language representation with \textbf{N}eighbor-aware \textbf{A}ttention), which is designed to better understand syntactic and contextual information based on comprehensive experimental evaluation of data processing, pre-training tasks, attention mechanisms. First, we extend the conventional pre-training tasks. Our models are trained to predict not only individual tokens but also an entire span of noun phrases during the pre-training procedure. This noun-phrase span masking scheme lets models learn contextualized representations in the whole span level, which benefits predicting answer texts for the specific extractive QA tasks. Second, we enhance the self-attention approach by incorporating a novel neighbor-aware mechanism in the transformer architecture~\citep{vaswani2017attention}. We find that more consideration of relationships between neighboring tokens by masking diagonality in attention matrix is helpful for contextualized representations. Additionally, we use a larger volume of corpora for pre-training language models and find that using a lot of additional datasets does not guarantee the best performance. 

We evaluate our proposed models on the SQuAD datasets which is a major extractive QA benchmarks for pre-trained language models. For SQuAD 1.1 task, ANNA achieves new state-of-the-art results of 90.6\% Exact Match (EM) and 95.7\% F1-score (F1). When evaluated on the SQuAD 2.0 development dataset, the results show that our proposed approaches obtain competitive performance outperforming self-supervised pre-training models such as BERT, ALBERT, RoBERTa, and XLNet models.

We summarize our main contributions as follows:
\begin{itemize}
\item We propose a new pre-trained language model, ANNA that is designed to address extractive QA tasks. ANNA is trained to predict the masked group of words that is an entire noun phrase, in order to better learn syntactic and contextual information by taking advantage of span-level representations. 
\item We introduce a novel transformer encoding mechanism stacking new neighbor-aware self-attention on an original self-attention in the transformer encoder block. The proposed method takes into account neighbor tokens more importantly than identical tokens during the computation of attention scores.
\item ANNA establishes new state-of-the-art results on the SQuAD 1.1 leaderboard and outperforms existing pre-trained language models for the SQuAD 2.0 dataset.
\end{itemize}

\section{Related works}
\paragraph{\emph{Pre-trained contextualized word representations\/}} There have been many recent efforts on pre-training language representation models aiming for capturing linguistic and contextual information, and the models have brought a significant improvement of performance in a variety of NLP tasks. ELMo~\citep{peters2018deep} is a deep contextualized word representation to learn complex characteristics of word use across linguistic contexts, and pre-trained models with these representations have shown noticeable improvements in many NLP challenges. BERT~\citep{devlin2018bert} is a pre-trained language model with a deep bidirectional long short-term memory, which learns context in text using the masked language modeling (MLM) and the next sentence prediction (NSP) objectives for self-supervised pre-training. The latest language models~\citep{liu2019roberta,lan2019albert,yang2019xlnet,radford2018improving,2019t5,lewis2019bart} influenced by BERT mainly employ the transformer architecture~\citep{vaswani2017attention} for pre-training but are trained with similar or extended to the pre-training objectives used in BERT implementation for enhancement of performance. There also exist many attempts to improve the capabilities of the standard transformer mechanism in contextualized word representations. 

\paragraph{\emph{Extension of MLM\/}} Many recent studies have attempted to use different pre-training objectives by extending the MLM task in language modeling including BART~\citep{lewis2019bart} and T5~\citep{raffel2019exploring}. ELECTRA~\citep{clark2020electra} introduces a new pre-training method of replaced token detection that replaces input tokens with alternative samples and detects whether the tokens are replaced or not. MASS~\citep{song2019mass} is pre-trained on the sequence to sequence framework where fragments of input sentences are masked, and the masked fragment is predicted in its decoder part. XLNet~\citep{yang2019xlnet} adopts a span-based masking approach that predicts a masked subsequent span of tokens in a context of tokens autoregressively. SpanBERT~\citep{joshi2020spanbert} and REALM~\citep{guu2020realm} employ a span masking scheme that masks spans of tokens rather than random individual tokens, and the model is designed to learn span representations during pre-training. Similarly, LUKE~\citep{yamada2020luke}, ERNIE~\citep{zhang2019ernie}, and KnowBERT~\citep{peters2019knowledge} learn joint representations of words and entities by incorporating knowledge of entity embeddings. 

\paragraph{\emph{Improvement of Attention Mechanism\/}} Since the standard transformer architecture has flexibility, many studies have shown the implementation of Transformer-based variants for improving further performance on language modeling and NLP tasks such as machine translation. \citep{shaw2018self} extends self-attention mechanism by incorporating embeddings of relative positions or distances between sequence elements, which is beneficial for performance improvement in machine translation tasks. \citep{yang2019context} introduces a context-aware self-attention approach that improves the self-attention with additional contextual information. \citep{sukhbaatar2019augmenting} presents a novel attention method extending the self-attention layer with persistent vectors storing information which plays a similar role as the feed-forward layer. \citep{fan2021mask} proposes a mask attention network that is a sequential layered structure incorporated a new dynamic mask attention layer with the self-attention and feed-forward networks.

\section{Methodology}

\begin{figure}[h]
\centering
\includegraphics[width=7cm, height=10cm]{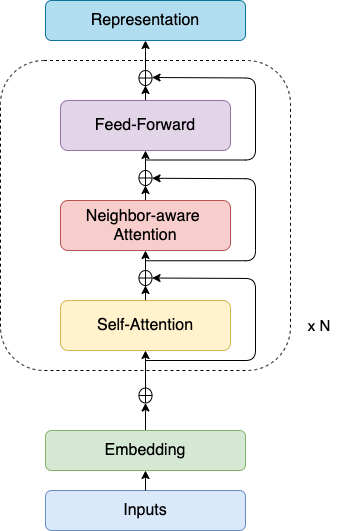}
\caption{Architecture of ANNA.}
\label{fig:fig1}
\end{figure}

We introduce a novel transformer encoder architecture integrating a new neighbor-aware mechanism for pre-training a language model. Figure \ref{fig:fig1} demonstrates the architecture of ANNA model. ANNA extends the original transformer encoder blocks by including a neighbor-aware self-attention layer stacked on a multi-head self-attention layer.

\begin{figure*}[t]
\centering
\includegraphics[width=2\columnwidth]{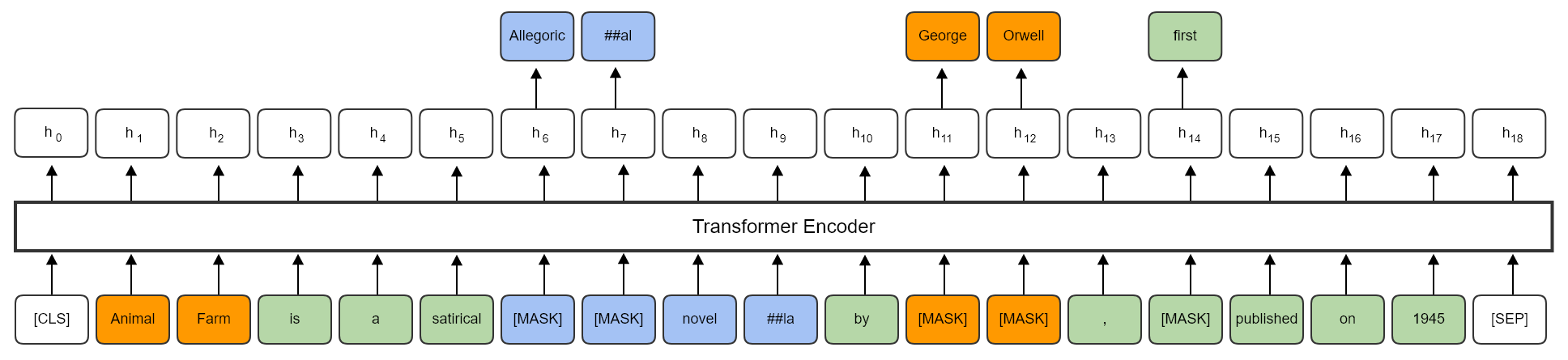}
\caption{Example of the input sequence ``\textit{Animal Farm is a satirical allegorical novella by George Orwell, first published on 1945}'' for pre-training ANNA. Different types of masking schemes are illustrated with such colors: masking a noun or noun phrase span (Orange), a whole word masking (Blue), and a wordpiece token masking (Green).}
\label{fig:fig2}
\end{figure*}

\subsection{Neighbor-aware Self-Attention}
\label{neighbor_aware_mechanism}
In this study, we propose a neighbor-aware attention mechanism. We assume that a single self-attention layer in the transformer encoder may be insufficient to learn context and the pre-trained models based on the transformer are hard to predict correct answers in downstream tasks due to linguistic noises brought in unrelated areas to a potential answer in the transformer encoder blocks. In an attention matrix, there is a pattern of diagonal line that illustrates a token more attends to itself, but less influences to other tokens. To give more attention to related tokens, we implement a new neighbor-aware attention mechanism that is designed to mitigate influences of identical tokens by ignoring the diagonality in an attention matrix when attention scores are computed. Instead, other tokens are more attended, so that the neighbor-aware mechanism enhances better understanding for relationships between tokens in inputs. Here, we integrate a neighbor-aware self-attention layer between the self-attention and the feed-forward network. The original attention information of a token, passed through the self-attention and the residual connection, is passed through the neighbor-aware self-attention again, so the token can more reflect a context to understand the sentence.

As the self-attention layer shown in Figure \ref{fig:fig1} is adopted from the standard transformer architecture ~\citep{vaswani2017attention}, we denote the self-attention as $A_{S}$ that is calculated using query (Q), key (K) and value (V) projections as follows:

\begin{equation}
  A_{S}(Q,K,V) = S_{S}(Q,K)V 
\end{equation}

\begin{equation}
  S_{S}(Q, K)=\Bigg[\frac{exp(Q_{i}K_{j}^{T} /\sqrt{d_{k}} )}{\sum_{k}exp(Q_{i}K_{k}^{T} / \sqrt{d_{k}})}\Bigg]
\end{equation}

where Q, K and V represent $HW_{q}$, $HW_{k}$ and $HW_{v}$, respectively. $H \in R\textsuperscript{$L\times$d}$ denoted as the input hidden vectors, L is the length of the input sequence, and d is the hidden size. $W_{q}, W_{k}, W_{v} \in R\textsuperscript{$d\times$d}$ are the projection matrices, and $d_{k}$ is the query/key dimension.
$A_{S}, A_{N} \in R\textsuperscript{$L\times$L}$ represents the attention matrices.

We define the Neighbor-aware Attention layer presented with $A_{N}$ as follows:

\begin{gather*}
  A_{N}(Q,K,V) = S_{N}(Q,K)V \\\\
  S_{N}(Q, K)=\frac{M(i,j)exp(Q_{i}K_{j}^{T} /\sqrt{d_{k}})}{\sum_{k}M(i,j)exp(Q_{i}K_{k}^{T} / \sqrt{d_{k}})}
\end{gather*}  
\begin{gather*}
  M(i,j) = 
  \begin{cases}
    0, & \mbox{if i = j }\\
    1, & \mbox{others}
  \end{cases}
\end{gather*}

where M denotes a mask that functions to omit capturing interactions of identical tokens. The interactions between each pair of input tokens $x_{i}$ and $x_{j}$ at positions \textit{i} and \textit{j} for 0 $\le$ \textit{i, j} $\le$ \textit{L} are calculated except for \textit{i} = \textit{j}.

\subsection{Pre-training Task}
We present a new pre-training task for training ANNA model. We follow the conventional MLM pre-training objective similar to BERT~\citep{devlin2018bert}. BERT is more sensible and effective to deeply represent context fusing the left and the right text with the MLM objective rather than unidirectional language models~\citep{radford2018improving,radford2019language,brown2020language} or shallow Bi-LSTM models~\citep{clark2018semi,huang2015bidirectional}. In addition, a new masking scheme is applied for focusing on noun phrases in order to train our language model for better understanding syntactic and lexical information considering the specific downstream tasks. Here, we define three different masking schemes as illustrated in Figure \ref{fig:fig2}. First, we use a span masking scheme that masks a group of texts in a span-level adopted by SpanBERT ~\citep{joshi2020spanbert}. In this study, nouns or noun phrases identified by spaCy's parser~\citep{spacy2} are randomly masked for span masking selection. Then we apply a whole word masking approach that masks all of the sub-tokens correspondings to a word at once, while we randomly mask tokens not included in the above two cases.

Following BERT, we randomly select 15\% of the tokens in input sequences, and 80\% of the selected tokens are replaced with the special token [MASK]. We keep 10\% of the tokens in the rest of them unchanged, and the other 10\% are replaced with randomly selected tokens. Our language model is also designed to train for the prediction of each token in the masked span by computing the cross-entropy loss function. However, the next sentence prediction (NSP) objective used in the BERT implementation is not used in this study, as RoBERTa~\citep{liu2019roberta} removes the NSP task due to performance decreases on downstream tasks.

\begin{table*}[ht]
\centering
\begin{tabular}{lclcl}
\hline Words &&  BERT tokens && ANNA tokens\\
\hline
Sant'Egidio && Sant , ' , E , \#\#gi , \#\#dio && Sant'Egidio \\
COVID-19 && CO , \#\#VI , \#\#D , - , '19' && COVID-19 \\
U.S. && U , . , S , . && U.S. \\
Ph.D. && Ph , . , D , . && Ph.D. \\
l'amour && l , ' , am , \#\#our && l'amour \\
non-profit && non , - , profit && non-profit \\
X-Files && X , - , Files && X-Files \\
UTF-16 && U , \#\#TF , - , 16 && UTF-16 \\
C++ && C , + , + && C++ \\
\hline
\end{tabular}
\caption{\label{table_tokenizer} Comparison of tokenization results between BERT and ANNA.}
\end{table*}

\subsection{Vocabulary and Tokenizer}
\label{vocab_tokenizer}
In this study, we build a new vocabulary of 127,490 wordpieces that are extracted from the English Common Crawl corpus~\citep{2019t5} and English Wikipedia dump datasets. The vocabulary consists of sub-words (30\%) tokenized by the WordPiece algorithm~\citep{wu2016google}, and 70\% of the rest include noun-phrase words in their original form. We aim to prevent words from being out of vocabulary words and also keep noun phrases as the original forms so that our model is able to take many words in order to better learn human linguistic understanding during training.

In addition, we propose a new approach of word tokenization to suit our vocabulary used to pre-train ANNA model. This approach avoids separating words by special symbols since our vocabulary contains words including special characters by tokenizing noun-phrase words with white space only. Many studies use a subword-based word representation method for efficiency in vocabulary. A word is represented with several subword units tokenized by BERT tokenizer as exampled in Table~\ref{table_tokenizer}. However, we do not follow this conventional tokenization method~\citep{wu2016google}, since we use a span masking scheme that masks an entire noun phrase randomly selected during a pre-training procedure. It is not suitable to train models as the length of masking tokens gets longer if subword units are used for the span masking scheme. We also aim to represent a whole-word token rather than subword units when attention scores are calculated. We implement an ANNA tokenizer in order to enhance a better understanding of contexts by not separating words as much as possible. Table ~\ref{table_tokenizer} compares word tokenization results between BERT and ANNA tokenizers.


\subsection{Pre-training Datasets} \label{pre-training_dataset}
We use an English Wikipedia dataset like BERT ~\citep{devlin2018bert}, and add publicly available English-language corpora such as a Colossal-Cleaned version of Common Crawl (C4) corpus ~\citep{2019t5}, Books3 ~\citep{gao2020pile}, and OpenWebText2 (OWT2) extended from WebText ~\citep{radford2019language} and OpenWebTextCorpus ~\citep{Gokaslan2019OpenWeb} for pre-training our models. Details of datasets and pre-processing techniques are described in Appendix~\ref{sec:appendix_2}.


With the extensive data pre-processing procedure, we gain the size of 12GB, 580GB, 51GB, and 22GB for Wikipedia, C4, Books3, and OWT2, respectively. The pre-processed texts are tokenized into 410B word-piece tokens in total for pre-training our models.

In this study, we conduct an experiment in order to investigate whether the use of different sources of data for pre-training language models affects model performance on downstream tasks. We compare the performance of models pre-trained with different datasets in Table ~\ref{table_corpora_performance_comparison}. We observe that C4 improves performance on the SQuAD 1.1 task when it is added to the Wikipedia dataset, but that models pre-trained over Books3 and OWT2 datasets are not beneficial for performance increases. We also find that the use of the larger volume of data including all of these four corpora is not helpful to improve performance. Thus we use both the C4 data and the Wikipedia corpus for pre-training ANNA models. Pre-training details for ANNA models can be found in Appendix A.

\begin{table}[h]
\centering
\begin{tabular}{l cc cc }
\hline Corpora && EM && F1 \\
\hline
Wikipedia && 85.51 && 90.99 \\
Wikipedia + C4 && \textbf{85.90} && \textbf{91.02} \\
Wikipedia + Books3 && 85.40 && 90.79 \\
Wikipedia + OWT2 && 84.79 && 90.27 \\
ALL && 85.14 && 90.22 \\
\hline
\end{tabular}
\caption{\label{table_corpora_performance_comparison} Comparison of model performance pre-trained with the different data sources. Models pre-trained with different pre-training corpora are evaluated on the SQuAD1.1 dataset. ALL includes the four datasets of Wikipedia, C4, Books3, and OWT2. Due to the limitation of computing resources, ANNA\textsubscript{Base} model is used for this experiment.}
\end{table}



\section{Experiments}
In this section, we present the fine-tuning results of ANNA transferred to specific extractive question answering tasks. 

We evaluate ANNA on SQuAD 1.1 and 2.0 tasks that are well-known machine reading comprehension benchmarks in the NLP area, and some NLU tasks. The dataset of SQuAD 1.1 consists of around 100k pairs of a question and an answer along with Wikipedia passages where the answers are included. This task is to predict a correct span of an answer text for a given question from the corresponding Wikipedia passage~\citep{rajpurkar2016squad}. For SQuAD 2.0, the dataset is extended to the SQuAD 1.1 dataset by combining over 50,000 unanswerable questions, so that systems are required to predict answers to both answerable and unanswerable questions~\citep{rajpurkar2018know}. We follow the fine-tuning procedure of BERT~\citep{devlin2018bert}, but the provided SQuAD training dataset only is used for fine-tuning, while BERT augments its training dataset with other QA datasets available in public. 

\begin{table*}[ht]
\centering
\begin{tabular}{ l c c c c } 
\hline
\multirow{2}{4em}{System} & \multicolumn{2}{c}{Dev} & \multicolumn{2}{c}{Test} \\
 & EM & F1 & EM & F1 \\
\hline
BERT\textsubscript{Large} ~\citep{devlin2018bert} & 84.2 & 91.1 & 85.1 & 91.8 \\
BERT\textsubscript{Large} (ensemble) & - & - & 87.4 & 93.1 \\
SpanBERT ~\citep{joshi2020spanbert} & - & - & 88.8 & 94.6 \\
XLNet\textsubscript{Large} ~\citep{yang2019xlnet} &  89.0 & 94.5 & 89.9 & 95.1 \\
LUKE ~\citep{yamada2020luke} & 89.8 & 95.0 & 90.2 & 95.4 \\
\hline
ANNA\textsubscript{Base} & 87.0 & 92.8 & - & - \\
\textbf{ANNA\textsubscript{Large}} & \textbf{90.0} & \textbf{95.4} & \textbf{90.6} & \textbf{95.7} \\
\hline
\end{tabular}
\caption{\label{table_squad_results} Performance of systems evaluated on the SQuAD 1.1 datasets.}
\end{table*}

\begin{table*}[h]
\centering
\begin{tabular}{l cc cc }
\hline \multirow{2}{6em}{System} && SQuAD 2.0 && SQuAD 2.0 \\
&& Dev EM && Dev F1 \\
\hline
BERT\textsubscript{Large}~\citep{devlin2018bert} && 79.0 && 81.8  \\
ALBERT\textsubscript{Large}~\citep{lan2019albert} && 85.1 && 88.1 \\
RoBERTa~\citep{liu2019roberta} && 86.5 && 89.4  \\
XLNet\textsubscript{Large}~\citep{yang2019xlnet} && 87.9 && 90.6  \\
ELECTRA\textsubscript{Large}~\citep{clark2020electra} && 88.0 && 90.6  \\
\hline
\textbf{ANNA\textsubscript{Large}} && \textbf{88.4} && \textbf{90.8} \\
\hline
\end{tabular}
\caption{\label{table_squad_2_results} Performance of systems evaluated on the SQuAD 2.0 development dataset.}
\end{table*}

\paragraph{SQuAD 1.1} Table ~\ref{table_squad_results} indicates the results of our best performing system compared with top results on the SQuAD 1.1 leaderboard. We also compare ours with BERT baselines. ANNA establishes a new state-of-the-art result on this task outperforming LUKE ~\citep{yamada2020luke} by EM 0.4 points and F1 0.3 points on the test dataset. LUKE is the latest best performing system in the leaderboard, and it is designed for contextualized representations of words and entities. As for a comparison with SpanBERT ~\citep{joshi2020spanbert} that masks contiguous sequences of token for span representations, ANNA also achieves better performance by both EM 1.8 points and F1 1.1 points.

\paragraph{SQuAD 2.0} ANNA is evaluated on SQuAD 2.0 development dataset, and the results are compared with the published pre-trained language models ~\citep{devlin2018bert,liu2019roberta,lan2019albert,yang2019xlnet,clark2020electra} in Table~\ref{table_squad_2_results}, which demonstrates that ANNA outperforms all of those language models and in particular, produces performance increases than ELECTRA by 0.4 points of EM and 0.2 points of F1.

\paragraph{GLUE} The General Language Understanding Evaluation (GLUE) benchmark is a collection of datasets used for training and evaluation diverse natural language understanding tasks~\citep{wang2018glue}. Since fine-tuning on GLUE is currently in progress, we show the results of the tasks that we complete in Appendix~\ref{sec:appendix_1}.

\section{Model Analysis}
We conduct additional experiments in terms of perspectives such as data processing, pre-training task, and attention mechanisms. We report a detailed analysis of how those approaches affect the performance of ANNA on a specific downstream task individually. In this study, ANNA\textsubscript{Base} model is used for these additional experiments due to the limitation of computing resources.

\subsection{Effect of ANNA Tokenization}
As mentioned in Section~\ref{vocab_tokenizer}, we build a new vocabulary containing noun-phrase words in their original format. For this, we introduce a new word tokenization strategy that keeps words in the original formats for noun phrases, which suits for our vocabulary. We compare our tokenization approach with the standard word-piece split approach, and find that ANNA tokenization performs better as shown in table~\ref{table_analysis_tokeniser}.

\begin{table}[h]
\begin{tabular}{l c c}
\hline \multirow{2}{6em}{} & SQuAD1.1 & SQuAD1.1 \\ & Dev EM & Dev F1 \\
\hline
WordPiece tokenizer  & 85.3 & 90.8 \\
ANNA tokenizer & \textbf{86.3} & \textbf{91.2} \\
\hline
\end{tabular}
\caption{\label{table_analysis_tokeniser} Ablation study of our tokenizer comparing to BERT tokenizer}
\end{table}

\subsection{Effect of Data Processing}
We describe several data pre-processing techniques we conduct to build a high-quality dataset for pre-training ANNA in Section \ref{pre-training_dataset}. Here we demonstrate how the use of the data processing techniques affects the performance on the extractive question answering task. There exist documents with a variety of ranges of word length in the pre-training corpora. For a generation of an input sequence, documents containing less than 100 words are filtered out, while the others are split into multiple sentence chunks. Due to the maximum sequence length of 512, we limit the size of the chunks to not exceeding approximately 300 words. We observe that the data processing procedure making a suitable word length for the max sequence length is helpful to improve performance slightly as shown in Table~\ref{table_analysis_data}. However, the input sequences overlapped with 128 tokens at the back and front between successive sentence chunks rather hurt system performance.   

\begin{table*}[h]
\centering
\begin{tabular}{l c c }
\hline \multirow{2}{8em}{Data Processing} & SQuAD1.1 & SQuAD1.1 \\
& Dev EM & Dev F1 \\
\hline
Wiki+C4  & \multirow{2}{3em}{85.9} & \multirow{2}{3em}{91.0} \\
(Without sentence chunking)  &  &  \\
Wiki+C4  & \multirow{2}{3em}{85.0} & \multirow{2}{3em}{90.5} \\
(Sentence chunking with 128 token-overlap) & & \\
\hline
Wiki+C4 & \multirow{2}{3em}{\textbf{86.3}} & \multirow{2}{3em}{\textbf{91.2}} \\
(Sentence chunking)  & & \\
\hline
\end{tabular}
\caption{\label{table_analysis_data} Comparison of model performance pre-trained with the use of different data processing techniques.}
\end{table*}

\subsection{Effect of Pre-training Mechanism}

\begin{table}[h]
\begin{tabular}{l cc cc }
\hline \multirow{2}{6em}{Model} && SQuAD1.1 && SQuAD1.1 \\
&& Dev EM && Dev F1 \\
\hline
Standard MLM && 83.7 && 89.1 \\
w/POS && 80.7 && 87.1 \\
Entity && 85.3 && 90.8 \\
\hline
Noun phrase && \textbf{86.3} && \textbf{91.2} \\
\hline
\end{tabular}
\caption{\label{table_analysis_masking} Results of different masking schemes during the pre-training task.}
\end{table}

We investigate how different MLM objectives affect the performance of models on a specific downstream task. During a pre-training procedure, a model is trained with a deep bidirectional representation of input sequences. First, we concatenate part-of-speech (POS) tags to each word, then we apply a whole word masking approach to explore whether a masking method employing syntactic information is helpful to understand the context. We also mask tokens identified as named entities and noun phrases instead of masking single tokens randomly. In all of the experiments, we use the same percentage of 15\% for the masking tasks. Table ~\ref{table_analysis_masking} compares results on the SQuAD 1.1 task for models using those MLM schemes. Comparing with the standard MLM approach that simply masks 15\% of tokens, the pre-trained models using Entity and Noun-phrase MLM schemes improve performance, but the approach masking words including POS tags decreases performance than the standard MLM. Thus we use the Noun-phrase MLM approach to pre-train ANNA models for final results. 

\subsection{Effect of Neighbor-aware Self-Attention}
We attempt to implement a new transformer encoder focusing on relatives, entities, or neighbors in input tokens in order to enhance capturing syntactic and contextual information. Firstly, we extend the original self-attention based on the transformer in order to consider relationships between input tokens. The relation matrix of input tokens is simply added when attention scores are computed. For an entity-self-attention that focuses on named entities, we identify named entities in text and then compute additional attention scores to those entities for learning effective representations. We describe the mechanism of a neighbor-aware self-attention in detail in Section~\ref{neighbor_aware_mechanism}. We report that the neighbor-aware self-attention approach performs better than the original self-attention and other transformer modifications on the extractive question-answering task in Table ~\ref{table_analysis_attention_type}. We consider that the neighbor-aware mechanism is effective to capture relation information of neighboring tokens in an input sequence.

\begin{table}[h]
\centering
\begin{tabular}{l c c }
\hline \multirow{2}{6em}{Model} & SQuAD1.1 & SQuAD1.1 \\
& Dev EM & Dev F1 \\
\hline
Self-Att. & 85.9 & 91.1 \\
Relative-QK-Att. & 86.0 & 91.1 \\
Relative-QV-Att. & 85.2 & 90.7 \\
Entity-Self-Att. & 85.7 & 90.9 \\
\hline
Neighbor-Aware-Att. & \textbf{86.4} & \textbf{91.4} \\
\hline
\end{tabular}
\caption{\label{table_analysis_attention_type} Comparison of model performance pre-trained with different transformer variants. Att is an abbreviation for Attention. The Self-Att. scores are the mean of multiple runs.}
\end{table}


\subsection{Effect of Layer-stacking Approach}
We examine how approaches to stack sub-layers in a transformer encoder architecture impact performance. We compose a transformer encoder block by collaborating three sub-layers such as a self-attention, a neighbor-aware self-attention, and a feed-forward network in different combinations. We evaluate the models using different combination methods of stacking layers and report the results on the SQuAD 1.1 dataset in Table~\ref{table_analysis_stack_attention}.

We observe that a self-attention substituted with a neighbor-aware attention in an original transformer architecture decreases performance by F1 0.5 points. When a neighbor-aware attention is stacked between a self-attention and a feed-forward network, the model slightly performs better than the original transformer. The sequential layered structure of a self-attention, a neighbor-aware attention, and a feed-forward network achieve the best performance on the exact matching criteria, which demonstrates that our proposed approach has an effect on the extractive question answering task. We consider that attention scores computed in a self-attention layer are re-weighted to actually related tokens by ignoring identical tokens during the computation of attention scores in the neighbor-aware attention so that the neighbor-aware mechanism is helpful to capture relationships between input tokens. 

\begin{table}[h]
\centering
\begin{tabular}{l c c }
\hline \multirow{2}{6em}{Model} & SQuAD1.1 & SQuAD1.1 \\ 
& Dev EM & Dev F1 \\
\hline
SA $\rightarrow$ FFN & 85.9 & 91.1 \\
NAA $\rightarrow$ FFN  & 85.5 & 90.6 \\
SA $\rightarrow$ SA $\rightarrow$ FFN  & 85.5 & 91.0 \\
NAA $\rightarrow$ NAA $\rightarrow$ FFN  & 86.1 & 91.5 \\
NAA $\rightarrow$ SA $\rightarrow$ FFN  & 86.1 & 91.4 \\  
\hline
SA $\rightarrow$ NAA $\rightarrow$ FFN  & \textbf{86.4} & \textbf{91.4} \\
\hline
\end{tabular}
\caption{\label{table_analysis_stack_attention} Performance of different stacking approaches of Self-attention (SA), Neighbor-aware-attention (NAA) and Feed-forward-network (FNN) layers in transformer encoder blocks. The SA-FNN scores are the mean of multiple runs.}
\end{table}


\section{Conclusion}
In this paper, we present a novel pre-trained language representation model, ANNA which improves the original transformer encoder architecture by collaborating a neighbor-aware mechanism, and is pre-trained for contextualized representations of words and noun phrases in a span level. The experimental results show that ANNA achieves a new state-of-the-art on the specific extractive question answering task by outperforming published language model systems including BERT baselines, as well as the latest top system on the corresponding leaderboard. There are two main directions for future research: (1) validating the competitiveness of ANNA to a variety of NLP tasks; and (2) enhancing the robustness of ANNA in order to apply for real-world question answering tasks in business.

\appendix
\section*{Appendix}
\label{sec:appendix}
\addcontentsline{toc}{section}{Appendices}
\renewcommand{\thesubsection}{\Alph{subsection}}

\subsection{Performance on GLUE }
\label{sec:appendix_1}
\begin{table*}[h]
\centering
\begin{tabular}{l c c c c c c c c c  }
\hline  & CoLA & SST-2 & MRPC & STS-B & QQP & MNLI & QNLI & RTE  & Avg.\\
\hline
BERT\textsubscript{Large} & 60.5 & 94.9 &  89.3/85.4 & 87.6/86.5 & 72.1/89.3 & 86.7/85.9 & 92.7 &  70.1  & 82.5  \\
SpanBERT & 64.3 & 94.8 &  90.9/87.9 & 89.9/89.1 & 71.9/89.5 & 88.1/87.7 & 94.3 & 79.0 & 85.0 \\
RoBERTa & 67.8 & 96.7 & 92.3/89.8 & 92.2/91.9 & 74.3/90.2 & 90.8/90.2 & 95.4 & 88.2 & 87.6 \\
\hline
ANNA & \textbf{65.8} & \textbf{96.4} &  \textbf{91.4}/\textbf{88.4} & \textbf{91.5}/\textbf{90.9} & \textbf{73.5}/\textbf{89.5} & \textbf{90.1}/\textbf{89.7} & \textbf{95.0} & \textbf{83.7} & \textbf{86.7} \\
\hline
\end{tabular}
\caption{\label{table_analysis_glue} Comparison results on the GLUE development set. The ``Avg.'' column is slightly different than the official GLUE scores, since the scores of WNLI and AX tasks are excluded in the average.}
\end{table*}

At this stage, we have not submitted our results to the official GLUE leaderboard~\footnote{https://gluebenchmark.com/leaderboard}, since we currently work on fine-tuning for the GLUE benchmark. Instead, we report our results on the tasks that we have completed the evaluation so far as shown in Table~\ref{table_analysis_glue}. We compare performance with two baseline models, BERT and SpanBERT, as the former is a pre-trained language model using a standard encoder architecture, and the later is pre-trained to predicts spans of texts, and motivated our noun-phrase masking approach. Comparing to the baselines, ANNA outperforms those baselines on every task, and gains the improvement of 1.7\% accuracy over SpanBERT in average. For further improvement of performance on GLUE, we continue to work on fine-tuning.

\subsection{Pre-training Datasets and Pre-processing}
\label{sec:appendix_2}
\begin{table*}[ht]
\centering
\begin{tabular}{l cc c c c }
\hline && Wikipedia & C4 & \textit{Books3} & \textit{OWT2} \\
\hline
Size of text && 16GB & 730GB & 100GB & 62GB\\
Token counts for text && 3.3B & 160B & 22B & 13B \\
\hline\hline
Size of pre-processed text && 12GB & 580GB & 51GB & 22GB\\
Token counts for pre-processed text  && 2.6B & 126B & 12B & 5B \\
\hline
\end{tabular}
\caption{\label{table_pretraining_corpora} Statistics of four corpora for pre-training including before and after the pre-processing procedure.}
\end{table*}

In this study, we use several large corpora for pre-training language models. As shown in Table ~\ref{table_pretraining_corpora}, the total size of data is about 900GB for the four corpora.

For pre-training language models with a large volume of corpora, it is crucial to generate high-quality data for inputs. We use heuristic pre-processing techniques to improve the data quality for the generation of input sequences as follows:
\begin{itemize}
    \item Each document is split into sentences, and we filter the sentences including less than 10 words out due to their incompleteness. Also, documents with less than 100 words are ignored for input sequences. 
    \item Text noises such as paragraph separators, special characters, URL addresses, and directory paths are heuristically filtered by regular expressions. 
    \item For Books3 data, non-English documents are deleted by a language-detection module ~\citep{shuyo2010language} which is utilized for the deletion of documents written in non-English words in the Common Crawl dataset.
    \item Since the maximum sequence length is 512 tokens, we split the pre-processed documents into multiple sentence chunks that do not exceed the predefined maximum length for the input of pre-training. 
\end{itemize}

\subsection{Pre-training Details}
\label{sec:appendix_3}
Table~\ref{table_hyperparameter} summarizes hyperparameters that we use for pre-training our two models: ANNA\textsubscript{Base} (L=12, H=768, A=12, Total Parameters=160M) and ANNA\textsubscript{Large} (L=24, H=1024, A=16, Total Parameters=550M). We use the maximum sequence length of 512, the Adam optimization ~\citep{kingma2014adam} with learning rates of 2e-4 and 1e-4 is used for the large and base models, respectively. 
Our large model ANNA\textsubscript{Large} is trained on 256 TPU v3 for 1M steps with the batch size of 2048, and it takes about 10 days. 

\begin{table}[h]
\centering
\begin{tabular}{l c c }
\hline Hyper-parameter & ANNA\textsubscript{Large} & ANNA\textsubscript{Base} \\
\hline
Number of layers & 24 & 12 \\
Hidden size & 1024 & 768 \\
FFN inner hidden size & 4096 & 3072 \\
Attention heads & 16 & 12 \\
Attention head size & 64 & 64 \\
Dropout & 0.1 & 0.1 \\
Warmup steps & 10k & 10k \\
Learning rates & 2e-4 & 1e-4 \\
Batch size & 2048 & 1024 \\
Weight decay & 0.01 & 0.01 \\
Max steps & 1M & 1M \\
Learning rate decay & Linear & Linear \\
Adam $\varepsilon$ & 1e-6 & 1e-6 \\
Adam $\beta_{1}$ & 0.9 & 0.9 \\
Adam $\beta_{2}$ & 0.999 & 0.999 \\
\hline
Number of TPU & 266 & 64 \\
Training time & 10 days & 5 days \\
\hline
\end{tabular}
\caption{\label{table_hyperparameter} Hyperparameters for pre-training ANNA models.}
\end{table}

\bibliography{acl}
\bibliographystyle{acl_natbib}

\end{document}